# Graph Colouring Problem Based on Discrete Imperialist Competitive Algorithm

Hojjat Emami[1] and Shahriar Lotfi[2]

[1]Department of Computer Engineering, Islamic Azad University, Miyandoab Branch, Miyandoab, Iran
`hojjatemami@yahoo.com`
[2]Department of Computer Science, University of Tabriz, Tabriz, Iran
`shlotfi@hotmail.com`

## ABSTRACT

*In graph theory, Graph Colouring Problem (GCP) is an assignment of colours to vertices of any given graph such that the colours on adjacent vertices are different. The GCP is known to be an optimization and NP-hard problem. Imperialist Competitive Algorithm (ICA) is a meta-heuristic optimization and stochastic search strategy which is inspired from socio-political phenomenon of imperialistic competition. The ICA contains two main operators: the assimilation and the imperialistic competition. The ICA has excellent capabilities such as high convergence rate and better global optimum achievement. In this research, a discrete version of ICA is proposed to deal with the solution of GCP. We call this algorithm as the DICA. The performance of the proposed method is compared with Genetic Algorithm (GA) on seven well-known graph colouring benchmarks. Experimental results demonstrate the superiority of the DICA for the benchmarks. This means DICA can produce optimal and valid solutions for different GCP instances.*

## 1. INTRODUCTION

Given an undirected and acyclic graph *G(V, E)*, a graph colouring involves assigning colours to each vertex of the graph such that any two adjacent vertices are assigned different colours. Graph Colouring Problem (GCP) is a complex and NP-hard problem [1, 2]. The smallest number of colours by which a graph can be coloured is called chromatic number. One of the main challenges in the GCP is to minimize the total number of colours used in colouring process. The GCP can be used to model problems in a wide variety of applications, such as frequency assignment, time-table scheduling, register allocation, bandwidth allocation, and circuit board testing [2-4]. So in applications that can be modelled as a GCP instance, it is adequate to find an optimal colouring of the graph. The GCP is NP-hard problem; therefore heuristic methods are suitable methods for solving the problem.

Imperialist Competitive Algorithm (ICA) is a stochastic search and optimization method which is inspired from imperialistic competition [5]. ICA has been used in many engineering and optimization applications. This algorithm is a population based algorithm i.e. instead of working with single solution, the ICA works with a number of solutions collectively known as population. Each individual in the population is called a country and can be either an imperialist or a colony. Colonies together imperialists form some empires. Movement of colonies toward their imperialists and imperialistic competition are the two main steps of the ICA. These operators hopefully causes the colonies converge to the global optimum of the problem. This algorithm has shown great efficiency in both convergence rate and better global optimum achievement [6].

The original ICA is inherently designed to solve continuous problems; therefore we did some changes in this algorithm and presented a discrete imperialist competitive algorithm (DICA). In this paper, we explore the application of DICA to solve the GCP and show this algorithm can find the valid solutions for this problem. Also in this paper the proposed method implemented and compared with genetic algorithm (GA). The experimental results on a variety of graph colouring benchmarks indicated the DICA method is efficient and superior to GA.

The rest of this paper is organized as follows. In Section 2 we briefly describe the theoretical foundation for this paper including graph colouring problem and its importance, description of GA and ICA techniques. In section 3 describes proposed discrete imperialist competitive algorithm and Section 4 illustrates how to solve the GCP by using DICA. Section 5 discusses on the results. Then, in Section 6, we briefly present some of the related works. Finally Section 7 draws some conclusion and gives an outlook of future works.

## 2. BACKGROUND

This section briefly describes graph coloring problem, imperialist competitive algorithm and genetic algorithm.

### 2.1. Graph Colouring Problem (GCP)

In graph theory the GCP is one of the most studied NP-hard problems. Many applications can be modelled by using the GCP such as scheduling [7], register allocation [8], frequency assignment, bandwidth allocation, and circuit board testing [2-4]. The GCP is an optimization problem that involves finding an optimal colouring for any given graph. Colouring of graph $G= (V, E)$ is a function $c: V \to C$, in which any two adjacent vertices $x, y \in V$ are assigned different colours, that is $\{x, y\} \in E \Rightarrow c(x) \neq c(y)$. $C$ is the set of all colours assigned to the vertices of graph. The function $c$ is called the colouring function that assigns colours to the vertices of graph. Optimal colouring for any given graph is one that uses exactly its predefined chromatic number. If we assume various assumptions in GCP there are many type of this problem. Generally there are two issues in graph colouring problem. One is that the graph vertices to be coloured correctly. In other words, all vertices of graph must be coloured and adjacent vertices have different colours. Another goal is that the total number of colours is minimized. In this paper we try to consider both goals.

To illustrate the process of colouring a graph, let us consider a graph $G= (V, E)$ as illustrated in Figure 1.a. This graph has 5 vertices and 5 edges (i.e. $|V| = 5\ and\ |E| = 5$). The chromatic number of this graph is 3 (i.e. $|K| = 3$). The coloured graph (one possible solution) indicated in Figure 1.b.

### 2.2. Genetic Algorithm (GA)

The GA is a well-known optimization and search algorithm which is inspired from evolution and natural genetics [9]. The GA has been applied to many science and practical applications [10]. The GA is a population based algorithm; this means instead of working with single solutions, it works with a set of solutions collectively known as a population. Like all evolutionary algorithms, a GA begins its work with an initial population. Each individual in this population is called a chromosome. Each chromosome must be assessed using a fitness function and assigned a goodness value to it. This fitness value is related to the objective function value of the problem.

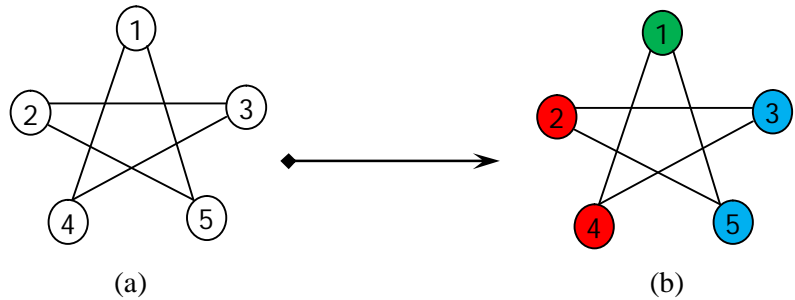

(a)                                 (b)

Figure 1. A simple example of graph colouring process.
(a) Graph *G* before colouring, (b) Graph *G* after colouring.

Selection operator among the population selects the best chromosomes and rejects the worst ones by using an appropriate selection rule. The output of the selection is an intermediate population. After selection operator is over, the intermediate population is updated using crossover and mutation operators to create the next population. In crossover two chromosomes are picked from the intermediate at random and some portions of chromosomes are exchanged between the chromosomes to create the new chromosomes. After crossover stage, mutation can occur. Mutation causes the GA escape from local optimums. A cycle of the selection, crossover and mutation creates one generation in GA. From one generation to the next, the population is updated until termination conditions are satisfied. A flowchart of GA is shown in Figure 2.

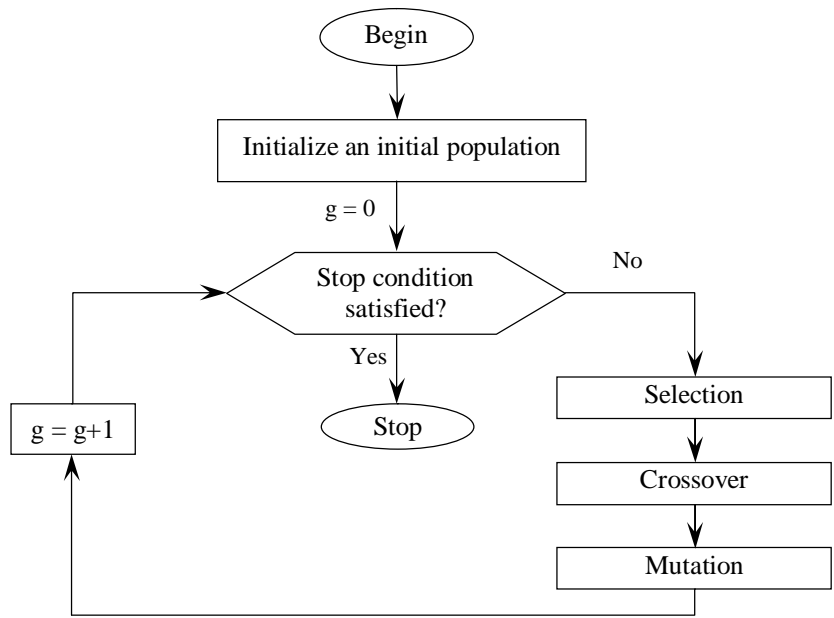

Figure 2. Flowchart of the GA technique

## 2.3. Imperialist Competitive Algorithm (ICA)

The ICA is one of the evolutionary population based optimization and search algorithms. The source of inspiration of this algorithm is the imperialistic competition. So far, the ICA has been used in various optimization and engineering applications [5, 6]. ICA has good performance in both convergence rate and better global optimum achievement. The ICA formulates the solution space of the problem as a search space. This means each point in the search space is a potential solution of the problem. The ICA aims to find the best points in the search space that satisfy the problem constraints. A flowchart of the working principle of the origin ICA is expressed in Figure 3.

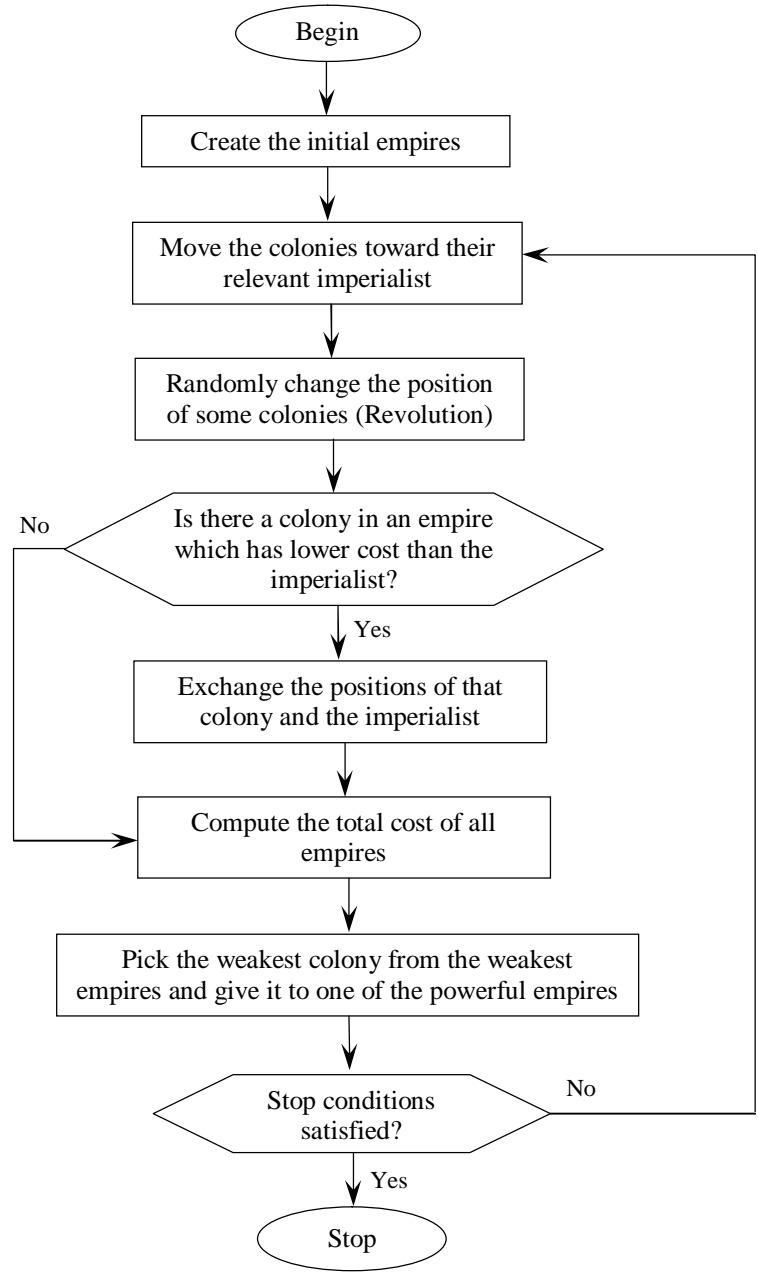

Figure 3. Flowchart of the Imperialist Competitive Algorithm

An ICA algorithm begins its search and optimization process with an initial population. Each individual in the population is called a country. Then the cost of each country is evaluated according to a predefined cost function. The cost values and their associated countries are ranked from lowest to highest cost. Some of the best countries are selected to be imperialist states and the remaining form the colonies of these imperialists. All colonies of the population are divided among the imperialists based on their power. Obviously more powerful imperialists will have the more colonies. The colonies together with their relevant imperialists form some empires. The ICA contains two main steps that are assimilation and imperialistic competition.

During assimilation step, colonies in each empire start moving toward their relevant imperialist and change their current positions. The assimilation policy causes the powerful empires are reinforced and the powerless ones are weakened. Then imperialistic competition occurs and all empires try to take the possession of colonies of other empires and control them. The imperialistic competition gradually brings about a decrease in the power of weaker empires and an increase in the power of more powerful empires. In the ICA, the imperialistic competition is modelled by just picking some of the weakest colonies of the weakest empire and making a competition among all empires to possess these colonies. The assimilation and imperialistic competition are performed until the predefined termination conditions are satisfied.

## 3. DISCRETE IMPERIALIST COMPETITIVE ALGORITHM (DICA)

This section describes a discrete version of imperialist competitive algorithm which is called DICA. The basic version of ICA is proposed to solve continuous problems. So with some modifications in some operators of the ICA, it can be used to solve discrete problems.

In the ICA, the assimilation operator causes colonies start moving to their relevant imperialists. The result of this process is to the colonies become more similar to their relevant imperialist states. Imperialist started to improve their colonies, on the other hand pursuing assimilation policy, the imperialists tried to absorb their colonies and make them a part of themselves. This operator must be changed to use in discrete problems. To model the assimilation policy in the discrete imperialist competitive algorithm, we used 2-point crossover. By using crossover, some random portion of imperialist and their relevant colonies are exchanged between them. In 2-point crossover operator, both countries (imperialist and a colony) are cut at two arbitrary place and the selected portion of both countries are swapped among themselves to create two new countries, as depicted is the following example.

**Example:** assume we want to colour a graph $G=(V,E)$, where $|V|=5$ and $|E|=5$. This graph is shown in Figure 1.a. Also suppose the following imperialist and colony countries. The cut points selected randomly and are $c_1 = 2$ and $c_2 = 3$. The new produced country is depicted below.

$$imperialist_i : \langle 1, \downarrow 2, 3, \downarrow 2, 1 \rangle \brace colony_i : \langle 3, \uparrow 1, 1, \uparrow 1, 2 \rangle \Rightarrow NewColony_i : \langle 3, 2, 3, 1, 2 \rangle$$

In the DICA, as the assimilation, the revolution operator needs to be changing too. Revolution operator causes a country suddenly change its position in the solution space. The revolution operator increases the exploration power of the ICA and helps it to escape from local optima. In the modified revolution, two different cells of a country are selected and then the selected cells are swapped among themselves. The revolution operator is illustrated in the below example.

**Example:** consider the below country be a candidate solution for the example graph illustrated in Figure 1.a. The new country after applying modified revolution is depicted as below.

$$colony_i : \langle 3, \vec{2}, 1, \bar{1}, 2 \rangle \Rightarrow NewColony_i : \langle 3, \bar{1}, 1, \vec{2}, 2 \rangle$$

## 4. APPLICATION OF DICA ON GRAPH COLOURING

This section describes how DICA is used to solve graph colouring problem. The input of the algorithm is an undirected and acyclic graph $G= (V, E)$, and the output is a reliable and optimal colouring for the input GCP instance.

At the start of procedure, a population of $N_{pop}$ countries is generated. If the GCP instance has $n$ vertices then each country is an array of $n$ colour indexes assigned to vertices of the graph. Figure 4.a illustrates a simple GCP instance that is to be coloured. This graph has 10 vertices, 15 edges, and its chromatic number is 3. Figure 4.b shows four countries created for the mentioned example graph. Each element of the countries is equivalent to a colour index. After creating initial population, the countries have to be assessed, according to the cost function expressed as follows:

$$\text{Cost(country)} = \begin{cases} \max_{i=1}^{N} & \text{if conflict} = 0 \\ \text{conflict} \times p + \max_{i=1}^{N} & \text{if conflict} \neq 0 \end{cases} \quad (1)$$

Where $p$ is the penalize coefficient and $N$ is the number of vertices of the graph. We compute how many unique colours are used in a country and the score for them specified by this number. Then some of the best countries are selected to be imperialists and the rest of the population forms the colonies of these imperialists. The imperialist states together with their colonies form some empires. Within the main iteration of the algorithm, imperialists try to attract their relevant colonies toward themselves and improve their cost. During this movement, if a colony reaches to a state that has smaller cost than its relevant imperialist, then they exchange their position. After assimilation, the imperialistic competition begins and all empires try to take the possession of colonies of other (weak) empires and control them. During this competition, the empires which are weaker than the others, loses their colonies. The outcome of this process is the extinction of the weakest empires. The DICA runs for a fixed number of replications, where a replication is defined as a cycle of assimilation, revolution, exchange, competition and elimination steps. Figure 5 summarizes the process of using discrete imperialist competitive algorithm on the graph coloring problem.

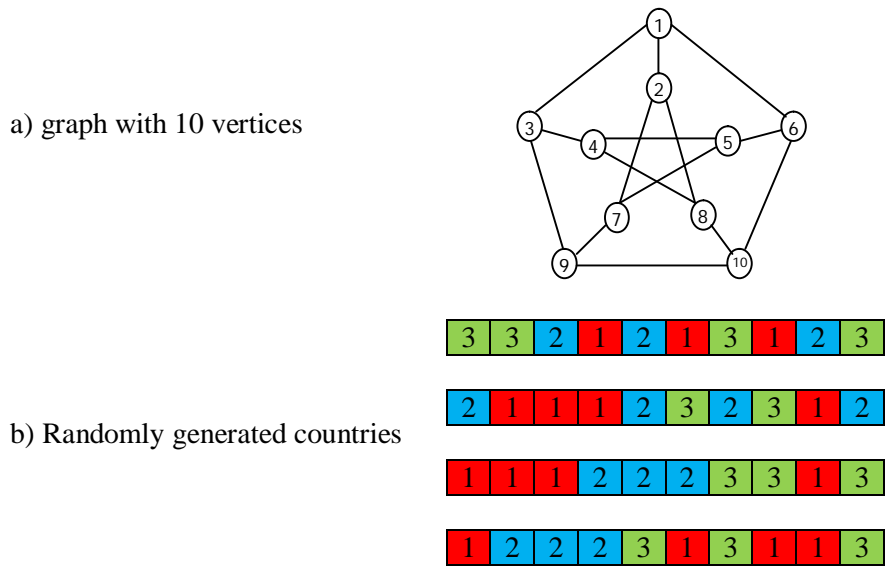

a) graph with 10 vertices

b) Randomly generated countries

Figure 4. An example graph and created random permuted countries

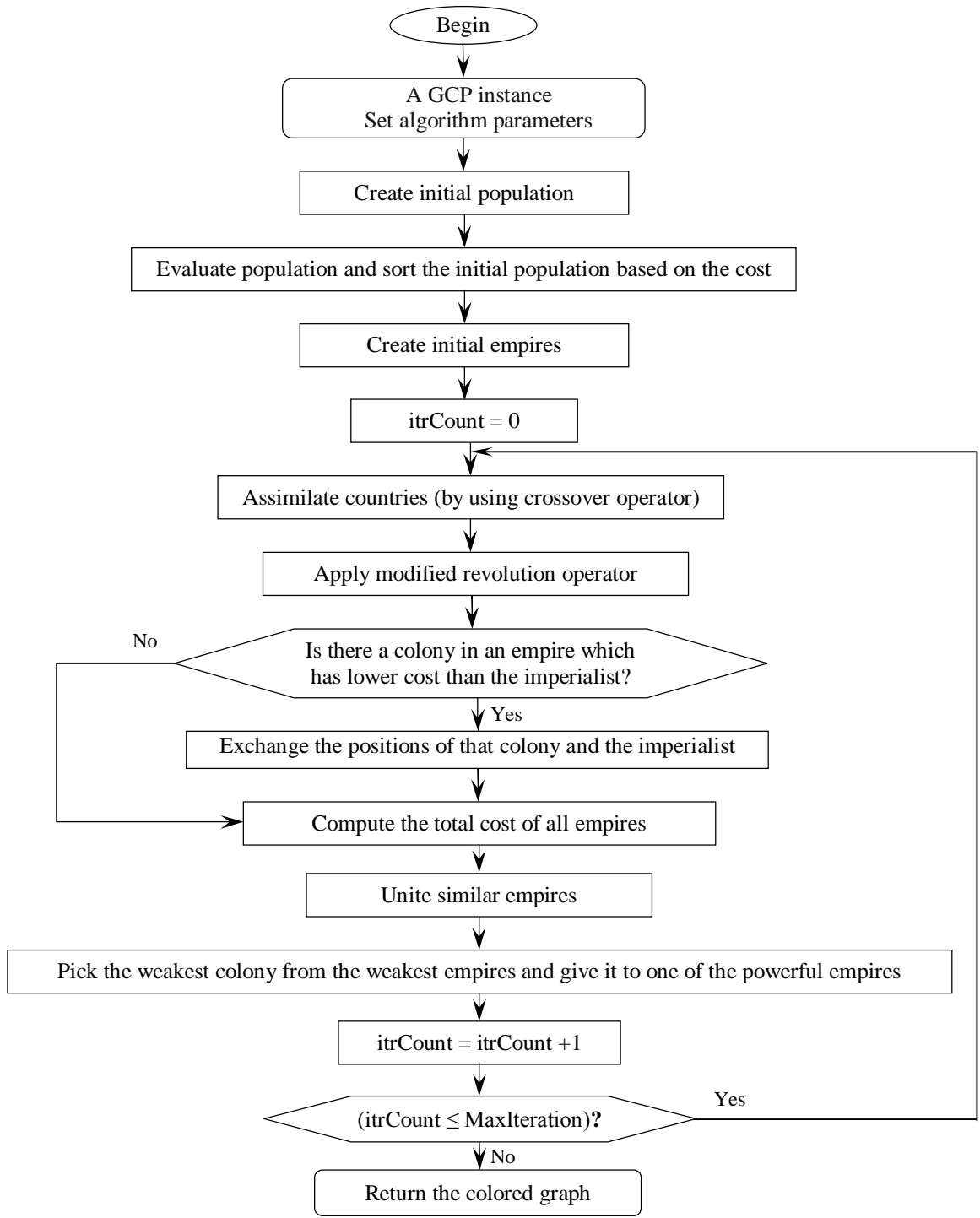

Figure 7. Flowchart of the process of applying discrete imperialist competitive algorithm on the graph colouring problem

# 5. EXPERIMENTAL RESULTS

In this section the efficiency of the proposed method is compared with GA on seven well-known graph colouring benchmarks. These benchmarks are Dataset1, Dataset2, Myceil3.col, Myceil4.col, Myceil5.col, queen5_5.col, and queen7-7.col. These data sets cover examples of data of low, medium and large dimensions. All data sets except Dataset1 and Dataset 2 are available at *http://mat.gsia.cmu.edu/COLOUR/instances*. Table 1 summarizes the characteristics of these benchmarks. Also Table 2 and 3 indicates the parameters set for DICA and GA in our implementations.

Table 1. Characteristics of data sets considered.

| Graph | Number of Vertices | Number of Edges | Chromatic Number |
|---|---|---|---|
| Dataset1 | 15 | 105 | 15 |
| Dataset2 | 20 | 190 | 20 |
| Myceil3.col | 11 | 20 | 4 |
| Myceil4.col | 23 | 71 | 5 |
| Myciel5.col | 47 | 236 | 6 |
| queen5_5.col | 25 | 160 | 5 |
| Queen7-7.col | 49 | 476 | 7 |

Table 2. The DICA algorithm parameters setup.

| Parameter | Value |
|---|---|
| Population size | 300 |
| Number of Initial Imperialists | 10 % of population size |
| Number of All Colonies | All population except imperialists |
| Number of Decades/ iteration count | 100 |
| Revolution Rate | 0.25 |
| Uniting Threshold | 0.02 |
| Assimilation Coefficient | 1.50 |
| Assimilation Angle Coefficient | 0.50 |
| Damp Ratio | 0.90 |

Table 3. The GA algorithm parameters setup.

| Parameter | Value |
|---|---|
| Population size | 300 |
| Mutation rate | 0.25 |
| Selection probability | 0.50 |
| Number of Generation / Iteration count | 100 |

## 5.1. Data Sets

Dataset1 is a complete graph which has 15 vertices and 105 edges. The chromatic number of this graph is 15. Dataset2 is another complete graph which has 20 vertices and 190 edges and its chromatic number is 20. Myceil3.col has 11 vertices and 20 edges. Myceil4.col has 23 vertices and 71 edges. Myceil5.col has 47 vertices and 236 edges. The chromatic number for Myceil3.col, Myceil4.col, and Myceil5.col are 4, 5, and 6 respectively. Queen5_5.col has 25 vertices and 160 edges. Queen7-7.col has 49 vertices and 476 edges. The chromatic number for Queen5-5 and Queen7-7 are 5 and 7 respectively.

## 5.2. Experimental Results

The algorithms are implemented using MATLAB software on a computer with 3.00 GHz CPU and 512MB RAM. In this section we evaluate and compare the performance of the DICA and GA algorithms on the graph colouring benchmarks. The efficiency of the DICA and GA algorithms is measured by the following criterion.

- The number of (success) failure over 20 runs of algorithm simulation.

How many the number of correct and successful runs will be higher then the efficiency of algorithm will be higher. Tables 4 shows the results (over 20 runs) obtained based on this measure. The results show the DICA method often works very well and finds the valid and optimal solution for different GCP instances. Also simulations show the size of population, the number of initial imperialist countries, the revolution rate, and also use an appropriate strategies for implementing the assimilation and revolution operators (in DICA) are effective to reach the optimal solutions. As mentioned in above sections, like to the mutation in the GA technique we selected a low revolution rate. For graphs that have few vertices we can use an initial population with fewer individuals and for high dimensional graphs we use a large initial population and also we can increase the number of iterations. In GA method, among different selection methods, we used roulette wheel to choose individuals to create next population. Also 2-pt crossover is used in the recombination process. Selection and mutation rate are 0.5, 0.3 respectively. For DICA the revolution rate and uniting threshold are set to 0.25 and 0.02 respectively.

As shown in Table 4, for Dataset1 the number of successful colourings of DICA and GA are same. For Dataset2 the number of successful iterations of DICA is greater than GA. The number of successful iterations of DICA for Myceil3.col, Myceil4.col and Myceil5.col data sets is greater than GA. Also the number of successful iterations of DICA for queen5-5.col and queen7-7.col are greater than GA. Simulation results indicate the runtime of DICA is lower than GA over on all data sets and this is due to the high convergence rate of the DICA method.

## 6. SUMMARY OF RELATED WORK

The GCP is one of the most important classical combinatorial optimization problems. So far, many researchers have been proposed different methods for solving the GCP. These methods fall into some broad categories such as polynomial-time approximation schemes, exact algorithms, greedy methods, parallel and distributed algorithms, decentralized algorithms, and heuristics [4, 11]. One of the most well-known methods in approximation schemes is the successive augmentation [4]. This method assigns a partial colouring to a small number of vertices and this process is extended vertex by vertex until the whole of graph is coloured.

Table 4. Results of DICA and GA algorithms on seven data sets; .he quality of solutions is evaluated using efficiency metric. The table shows success (failures) for 10 independent runs.

| Graph | Number of Vertices | Number of Edges | DICA Success (*Failure*) | GA Success (*Failure*) |
|---|---|---|---|---|
| Dataset1 | 15 | 105 | 20*(0)* | 20*(0)* |
| Dataset2 | 20 | 190 | 19*(1)* | 18*(2)* |
| Myciel3.col | 11 | 20 | 20*(0)* | 20*(0)* |
| Myciel4.col | 23 | 71 | 20*(0)* | 18*(2)* |
| Myciel5.col | 47 | 236 | 18*(2)* | 17*(3)* |
| queen5_5.col | 25 | 160 | 18*(2)* | 16*(3)* |
| queen7_7.col | 49 | 952 | 17*(3)* | 15*(5)* |

Algorithms for finding optimal colourings are frequently based on implicit enumeration [4]. Brute-force search technique is one of the best well-known exact colouring methods [11]. In these techniques all solutions are checked for finding a reliable and optimal colouring for a graph and have high runtime. In the greedy algorithms, vertices of the graph are coloured in a specific order. The two best examples of greedy algorithms are DSATUR and (Recursive Largest First) RLF [12]. NP complete problems can easily be solved by using distributed computing and parallelism. In the distributed algorithms, graph colouring problem is related to the symmetry breaking problem. Randomized algorithms are faster methods for large dimensional graphs. The fastest algorithm in the class of randomized algorithms is the method presented by Schneider et al. [13].

Since graph colouring problem is an NP-hard problem, several artificial intelligence techniques have been applied on graph colouring problem including algorithms based on neural networks [14], DNA parallel approach (e.g. in [15]), learning automata (e.g. in [16]), evolutionary algorithms, hybrid methods (e.g. in [17] and [18]), scatter search [19], and local search algorithms (e.g. Tabu search [20] or simulated annealing [21]).

Since our work deals with finding optimal solutions for graph colouring by using an evolutionary algorithm, we discuss previous work on only some recently evolutionary algorithms that used for the GCP in detail.

Anh et al. presented an approach to the GCP using PSO algorithm that improves a simple deterministic greedy algorithm [22]. They proved that their proposed method is better than known heuristic algorithms. Lixia and Zhanli proposed a novel bi-objective genetic algorithm which employs an effective crossover and simple mutation as the genetic operators [23]. The authors claimed that their method is a promising approach to the GCP. Authors in [24] presented a hybrid chaotic ant swarm approach for the GCP which is called CASCOL. This approach is based on a chaotic ant swarm (CAS) and a simple greedy sequential colouring, first-fit algorithm. Their experimental results indicate that the proposed method is an efficient and competitive algorithm. A max-min ant approach is presented by Mohamed and Elbernoussi for the sum colouring problem which is an extension of ant system and a local heuristic [25]. Sum colouring problem is derived from the GCP. This approach aims to minimize the sum of colours that used to colour the graph. Fister and Brest was developed an approach based on differential evolution for graph colouring [26]. They compared their proposed method with some of the best heuristics and by extensive experiments showed their method is an efficient approach for graph colouring. Dorrigiv and Markib used artificial bee colony (ABC) algorithm to resolve graph colouring problem [27]. The proposed method is called ABC-GCP and its performance is evaluated based on the randomly generated graphs with different densities. Experimental results showed this method is a capable algorithm compared to other methods. A hybrid multi-objective genetic algorithm for bandwidth multi-colouring problem is presented in [28. Bandwidth multi-colouring is an extension of the GCP. In the proposed method, genetic operators are replaced with new ones which appropriate to the structure of the problem. It seems this method be better than the other standard genetic algorithm in solving GCP. These are only some of the proposed methods based on the evolutionary optimization algorithms for the graph colouring. Nonetheless graph colouring problem is an active research topic.

## 7. CONCLUSIONS

In this paper, we have presented an approach to graph colouring problem based on discrete imperialist competitive algorithm. The experiment is performed on seven graph colouring benchmarks. From the numerical simulation results, it can be concluded that the proposed method has enough power in solving different graph colouring problem instances. Discrete imperialist competitive algorithm needs less runtime to achieve the global optimums while genetic algorithm requires more runtime to achieve a solution. The proposed method can use for both low and high dimension graphs and can find optimal solutions for them. The proposed method can combine with other evolutionary or classic methods to find optimal solutions to graph colouring problem. One drawback of proposed method is that it may not find the optimal solutions in some times and this problem returns to the nature of evolutionary algorithms. In our future work, we will focus on presenting new evolutionary algorithm to solve graph colouring problem that have high efficiency compared to other available models.